# Investigating the Effect of CPT in Lateral Spreading Prediction Using Explainable AI

Cheng-Hsi Hsiao,[1] Ellen M. Rathje, Ph.D.,[2] and Krishna Kumar, Ph.D.[3]

[1]Maseeh Department of Civil, Architectural and Environmental Engineering,
University of Texas at Austin, Austin, TX, United States; E-mail: chhsiao@utexas.edu
[2]Maseeh Department of Civil, Architectural and Environmental Engineering,
University of Texas at Austin, Austin, TX, United States; E-mail: e.rathje@mail.utexas.edu
[3]Maseeh Department of Civil, Architectural and Environmental Engineering,
University of Texas at Austin, Austin, TX, United States; E-mail: krishnak@utexas.edu

## ABSTRACT

This study proposes an autoencoder approach to extract latent features from cone penetration test profiles to evaluate the potential of incorporating CPT data in an AI model. We employ autoencoders to compress 200 CPT profiles of soil behavior type index ($I_c$) and normalized cone resistance ($q_{c1Ncs}$) into ten latent features while preserving critical information. We then utilize the extracted latent features with site parameters to train XGBoost models for predicting lateral spreading occurrences in the 2011 Christchurch earthquake. Models using the latent CPT features outperformed models with conventional CPT metrics or no CPT data, achieving over 83% accuracy. Explainable AI revealed the most crucial latent feature corresponding to soil behavior between 1-3 meter depths, highlighting this depth range's criticality for liquefaction evaluation. The autoencoder approach provides an automated technique for condensing CPT profiles into informative latent features for machine-learning liquefaction models.

## INTRODUCTION

Lateral spreading is a particularly devastating consequence of liquefaction induced by earthquakes, wherein saturated soils experience significant permanent ground deformations toward a free face, such as a slope or riverbank. During an earthquake, the dynamic shaking generates excess pore water pressure in the saturated soil, causing a temporary loss of shear strength, effectively liquefying the soil. This liquefied soil behaves like a dense liquid, allowing the soil mass to deform laterally due to gravitational and inertial forces, particularly in areas adjacent to free faces. The lateral spreading can result in ground cracking, differential settlements, and the lateral displacement of structures, foundations, and buried utilities, leading to extensive damage and financial losses. Evaluating the potential for lateral spreading is crucial in liquefaction hazard assessment, particularly in areas with gently sloping topography and free faces.

Liquefaction triggering has been extensively studied to establish a link between the occurrence of liquefaction and soil characteristics. Soil properties are commonly investigated using the Standard Penetration Test (SPT) and Cone Penetration Test (CPT). The SPT-based triggering procedures (Seed et al. 1984, Youd et al. 2001, Idriss and Boulanger 2004) utilize $(N1)_{60cs}$, while the CPT-based procedures (Robertson and Wride 1997, Idriss and Boulanger 2004) use $q_{c1n}$. With the triggering model, engineers can examine the liquefaction at every profile data point. However, liquefaction occurrence within a soil profile does not necessarily imply a surface manifestation of liquefaction.

Ishihara (1985) proposed that liquefaction manifestation relates to the thickness of the overlying non-liquefiable layer (H1) and the underlying liquefiable layer (H2) based on SPT blow counts and groundwater table depth. Iwasaki (1984) used the factor of safety (FS) based on the triggering model to compute the liquefaction potential index (LPI) with a weight function to emphasize the importance of the top layer properties. Nevertheless, traditional methods for evaluating liquefaction manifestation have limitations in accuracy due to the complexity of soil response subjected to dynamic motion. Consequently, researchers have developed various machine learning (ML) methods to tackle the intricacies of this problem.

In recent years, ML-based liquefaction evaluation has gained popularity due to the high accuracy of these models. Most models utilize site parameters (e.g., PGA, GWT, and $M_w$) and soil characteristics (e.g., SPT-N, $q_c$, $\sigma_v$, $V_s$ and FC) of the "critical layer" as input parameters (Zhao et al., 2021; Zhou et al., 2022; Demir & Şahin, 2022). However, the issue with this approach is that the "critical layer" is defined based on engineering judgment and is assumed to represent the properties of the entire soil profile.

Some studies have proposed extracting information from the soil profile for ML model inputs. Durante & Rathje (2021) computed the median and standard deviation of the cone penetration test (CPT) profile 4 meters below the groundwater table as inputs for lateral spreading models. However, they reported that including extracted features from CPT profiles did not improve the accuracy. They concluded that the chosen CPT features did not adequately represent the soil profiles.

Therefore, this study explores how to extract features that represent detailed information from CPT data, which can be used as soil condition parameters for the downstream ML models. Notably, we propose a CPT feature extraction approach that is independent of engineering judgment, eliminating the need for subjective interpretations.

Feature extraction, or dimension reduction, is crucial before training an ML model. Raw data might carry redundant information, which can cause the model to overfit. One of the most common approaches to reduce dimensionality is Principal Component Analysis (PCA; Pearson, 1901). PCA computes principal components (axes) that maximize the variance of the original data. It projects the original features into a latent feature space, where the latent axes are orthogonal. The latent features are a linear combination of the original features. However, a downside of PCA is that it does not consider the order of the original features, which contradicts the importance of order in soil profiles. Moreover, due to the nature of PCA, the latent features cannot capture non-linear relationships among the original features.

To address these problems, this study proposes an autoencoder (Hinton & Salakhutdinov, 2006) for extracting features from CPT profiles. An autoencoder is an unsupervised technique commonly used in image and natural language processing (NLP). The original features pass through a bottleneck-like neural network in the autoencoder. The narrowest part of the network is where the latent features are extracted. Due to the complexity of neural networks, the autoencoder's latent features can describe the non-linear behavior of the original features. Additionally, positional encoding is implemented, which allows the autoencoder to consider the original features as a sequence of inputs. Thus preserving the information order of the soil profile. The methodology section will explain the details of the autoencoder and positional encoding.

We will incorporate the latent features of the CPT profiles collected after the 2011 Christchurch lateral spreading event (Durante & Rathje, 2021) to train XGBoost models (Chen & Guestrin, 2016) for lateral spreading predictions. Although these latent features do not have a direct physical meaning, we use a sampling approach to visualize the representation of latent feature. Lastly, we will explain latent features using SHapley Additive exPlanations (SHAP; Lundberg and Lee, 2017) approach.

**DATA**

We collect 12,000 sites of CPT profiles from the New Zealand Geotechnical Database (NZGD) to train the autoencoder model. Figure 1 shows the region of the study area and the distribution of collected CPT sites (black dots). We only choose the CPT sites with a penetration depth of over 10 m. We compute the Ic and qc1Ncs from the raw CPT data based on Idriss and Boulanger (2008) procedure. Due to the different sample rates of CPT data (e.g., 1 cm or 2 cm), we regularize the interval by taking the average over every 5 cm of the 10 m profile. The resulting $I_c$ and $q_{c1Ncs}$ profiles are arrays with a length of 200, which are later used for feature extraction.

In addition to the CPT data, we use Durante & Rathje's (2021a) lateral spreading dataset to evaluate the extracted CPT features from this study. The lateral spreading dataset includes 6,704 data points (see yellow circles in Figure 1). Each data point has five site parameters, peak ground acceleration (PGA), groundwater depth (GWD), the distance to the rivers (L), slope, and elevation, for the 2011 Christchurch Earthquake.

In this study, we will train an autoencoder with collected CPT sites. The training-validation-testing split for the CPT profiles is 70:15:15. The trained autoencoder will generate latent features for Ic and qc1Ncs profiles. We will use the CPT latent features incorporated with site parameters from Durant and Rathje's dataset to predict the lateral spreading occurrence along the Avon River. Since we use site parameters from their dataset and latent features from our CPT data, only 3,364 data points are in both datasets. The original Durante & Rathje (2021b) dataset is available in DesignSafe (Rathje et al. 2017). The scripts and developed models are available on GitHub (https://github.com/chhsiao93/xai-cpt).

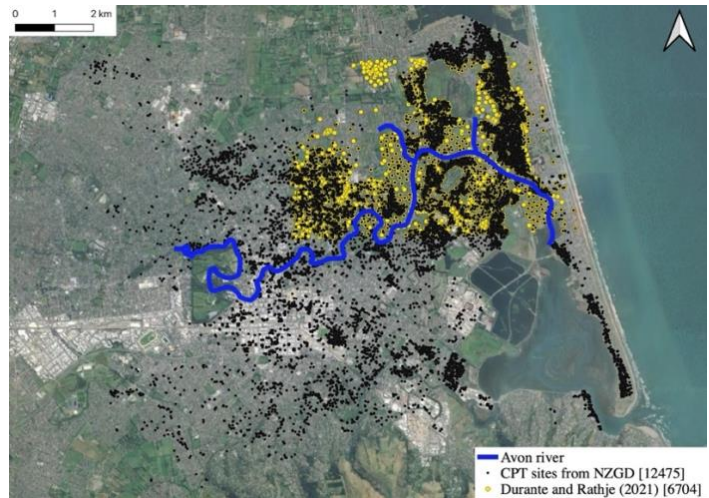

Figure 1. The spatial distribution of CPT sites in Christchurch region.

**METHODOLOGY**

*Autoencoder*

An autoencoder (Hinton & Salakhutdinov, 2006) is a neural network used for unsupervised learning, particularly in dimensionality reduction and feature extraction tasks. It comprises two main components: an encoder and a decoder (see Figure 2). The encoder compresses the input data into a lower-dimensional representation, i.e., a latent space capturing the most critical features while discarding redundant information. The decoder then reconstructs the original data from this compressed form. The autoencoder network is trained to minimize the difference between the input data and its reconstruction. In our study, the autoencoder is trained to reduce the mean square error (MSE) between the input CPT profile and the reconstructed profile.

The dimension of the latent features is a hyperparameter that highly impacts an autoencoder's ability to reconstruct data. Fewer latent features typically result in poorer reconstruction quality. While increasing the number of latent features improves reconstruction, it contradicts the goal of dimensionality reduction. In this study, we compress 200 data points from a CPT profile into 10 latent features, achieving a 20-fold reduction in dimensionality while maintaining a good level of reconstruction.

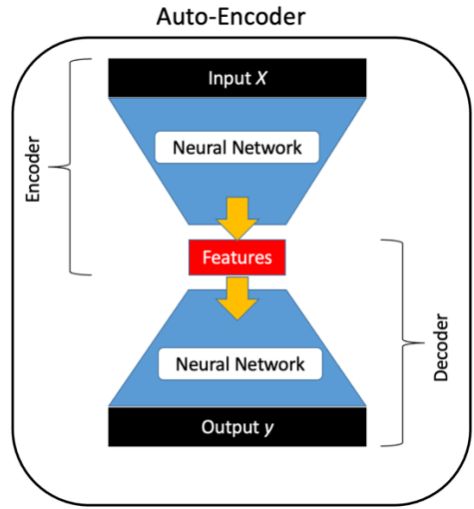

Figure 2. The structure of an autoencoder.

*Positional Encoding*

We should consider any CPT profile as a sequential input with crucial positional information. For instance, a 1-meter clay layer significantly decreases the risk of liquefaction if it is at the top of the profile but has a lesser impact at a deeper depth. We implement positional encoding into our CPT data to account for positional information.

Positional encoding is an approach to inject positional information before the autoencoder. The positional encoding for each position pos and each dimension I of the embedding is defined as follows:

$$PE_{(pos,2i)} = \sin\left(\frac{pos}{10000^{2i/d}}\right) \quad \text{Eq. 1}$$
$$PE_{(pos,2i+1)} = \cos\left(\frac{pos}{10000^{2i/d}}\right) \quad \text{Eq. 2}$$

Where *pos* is the position representing the input sequence, *i* is the dimension index, and *d* is the total dimension of the model's embeddings.

Let's consider a sequential input with 10 sequences, each having 20 feature dimensions. We apply positional encoding to this input using Eq. 1 and Eq. 2. In Figure 3a, the rows represent the input sequences, the columns represent the features of each sequence, and the color indicates the positional encoding output. Positional encoding generates unique signals for each sequence: adjacent sequences have similar signals, while distant sequences have distinct ones. For example, the positional signal for Sequence 0 (the first row) is similar to that for Sequence 1 (the second row) but differs from that for Sequence 9 (the last row). The positional encoding is added to the input profile, giving each sequence a unique representation in the feature domain. This process is embedded at the beginning of the encoder and does not interfere with the autoencoder's goal of minimizing the difference between the input and output.

We want to apply the positional encoding to CPT data. However, the feature dimension is too few (we only have $I_c$ and $q_{c1Ncs}$ measurements at each depth) to generate a unique positional signal. To compensate for this, we reshape the CPT profile (either $I_c$ or $q_{c1Ncs}$) from a 1x200 array to a 10x20 matrix, where each row contains a 1-meter profile, and each element represents a 5-cm average CPT data point. This reshaping transforms the profile into ten sequential inputs, each with 20 features (see Figure 3b).

In this study, we train two autoencoders for $I_c$ and $q_{c1Ncs}$ profiles. Each autoencoder will compress the profile from a 10x20 matrix to 10 features and reconstruct the profile from these ten features. We will use the extracted ten features to represent the profile in latent space.

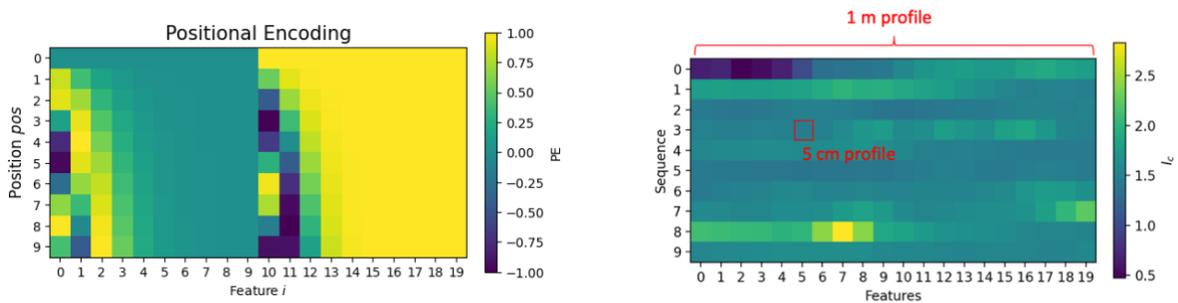

a.　　　Positional encoding　　　　　　　　　　b.　Reshaped $I_c$ profile

Figure 3. Inputs for autoencoder

**RESULTS**

*Reconstruction of CPT Profile*

We develop two autoencoders for extracting latent features from $I_c$ and $q_{c1Ncs}$ profiles. To better understand the reconstruction error, we plot samples of $I_c$ and $q_{c1Ncs}$ profiles from the testing data. We select three samples representing small, medium, and large errors and show them in Figure 4. The figure shows the profiles with minor errors, which are straight lines or smooth curves. The more the "noise" in the profile, the larger the autoencoders would generate the errors. The autoencoders can capture the general trend of the profiles but fail to describe the drastic change. This problem indicates that the information of the original profile is overcompressed: 10 latent features are not enough to compress all the details. One of the solutions is to increase the number of latent features. However, increasing the number of latent features and dimension reduction somehow conflict. Therefore, this study uses ten latent features in the following work. Researchers can decide how many latent features to use depending on their purpose and how well they want the reconstructions.

To evaluate the autoencoder, we measure the reconstruction error, which is the difference between the reconstructed and original profiles. We measure the reconstruction error using two metrics: RMSE and log difference. The calculations of two metrics are shown in Eq. 3 and Eq. 4, respectively.

$$RMSE = \sqrt{\frac{\sum_{i=1}^{n}(y_i - X_i)^2}{n}} \qquad \text{Eq. 3}$$

$$Abs\ Log\ Difference = \frac{\sum_{i=1}^{n}|\log y_i - \log X_i|}{n} \qquad \text{Eq. 4}$$

$X_i$ is i$^{th}$ data point of the original profile, $y_i$ is the i$^{th}$ data point of the reconstructed profile, $n$ is total number of points of a profile (200 for this study).

RMSE measures the average Euclidean distance between actual values and reconstructed values; Log difference measures the logarithm of the ratio of reconstructed values to actual values. Figure 5 shows the reconstruction error of $I_c$ and $q_{c1Ncs}$ of testing data regarding RMSE and absolute log difference. The results show that the reconstructed $I_c$ has a range of RMSE between 0.05 to 0.4 and a range of absolute log difference between 0.025 and 0.15. On the other hand, the reconstructed $q_{c1ncs}$ has a range of RMSE between 5 and 30 and a range of absolute log difference between 0.05 and 0.18.

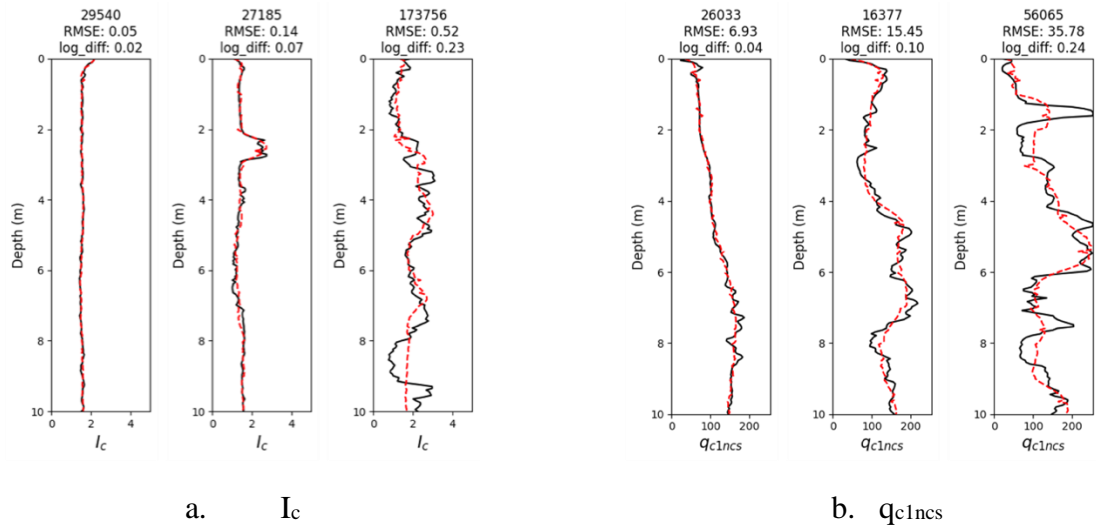

a. $I_c$                                           b. $q_{c1ncs}$

Figure 4. Reconstruction samples from testing data

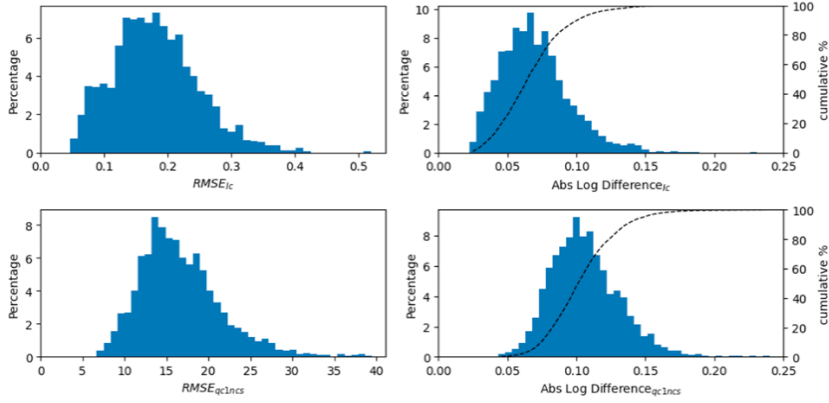

Figure 5. Reconstruction error of autoencoder of testing data

*Predictive Performance on Lateral Spreading Prediction*

To evaluate the effectiveness of the extracted latent features, we developed four XGBoost models to predict lateral spreading occurrence. Each model uses different combinations of features as inputs. Table 1 summarizes the input features for each model, where the number in parentheses indicates the number of features in a feature group, and the symbol '✓' indicates the inclusion of that feature group in a model.

Table 1. Input features and model configuration

|  |  | A | B | C | D |
|---|---|---|---|---|---|
| Feature Group | Site parameters (5) | ✓ | ✓ | ✓ | ✓ |
|  | STD and median of 4-m CPT (4) |  | ✓ |  |  |
|  | 1-m average CPT (20) |  |  | ✓ |  |
|  | CPT latent (20) |  |  |  | ✓ |
| Configuration | Max depth | \multicolumn{4}{c}{11} |
|  | Early-stopping rounds | \multicolumn{4}{c}{5} |
|  | No. estimators | 13 | 13 | 18 | 22 |

Table 1 also details the configurations used for training the XGBoost models. The "max depth" parameter specifies the maximum depth a decision tree can grow in XGBoost. The "number of estimators" parameter also indicates the number of trees in the model. To optimize each model, we employed early stopping techniques. Early stopping halts the training process if the model's performance on validation data does not improve for a specified number of rounds. In this study, we set the early stopping round to 5. This means that the training process will stop if validation accuracy does not improve for five consecutive rounds, ensuring that the final model is the best-performing one before any performance plateau.

We evaluate the optimized models using testing data. Figure 6 displays the confusion matrices for each model, illustrating the distribution of predictions across the categories. For instance, Model A correctly predicts 220 'No lateral spreading' cases, which are True Negatives (TNs), and 181 'Yes lateral spreading' cases, which are True Positives (TPs). However, it incorrectly predicts 53 'No' cases as 'Yes' (False Positives, FPs) and 51 'Yes' cases as 'No' (False Negatives, FNs).

Figure 6 shows that Model B has the highest number of True Positives (TPs) with 188, while Model D has the highest number of True Negatives (TNs) with 242. Despite these differences, there are no significant variations among the confusion matrices of the models.

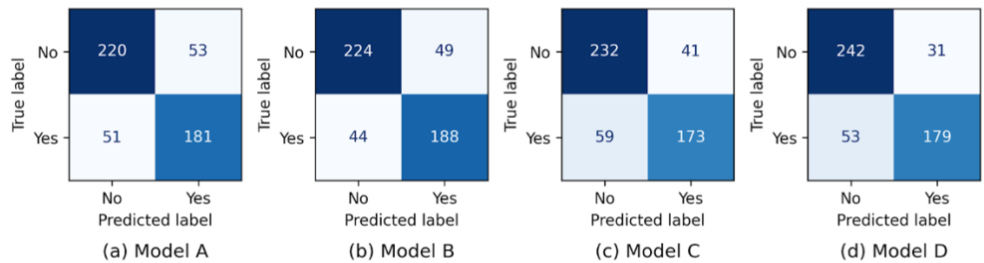

Figure 6. Confusion matrix of testing data

Table 2. Model Performance on Testing Data

| Metric/Model | A | B | C | D |
|---|---|---|---|---|
| $Accuracy = \dfrac{TP + TN}{Total\ Instances}$ | **0.79** | 0.82 | 0.80 | **0.83** |
| $Balanced\ Accurac = \dfrac{1}{2}\left(\dfrac{TP}{TP + FN} + \dfrac{TN}{TN + FP}\right)$ | **0.79** | 0.82 | 0.80 | **0.83** |
| $Precision = \dfrac{TP}{TP + FP}$ | **0.77** | 0.79 | 0.81 | **0.85** |
| $Recall = \dfrac{TP}{TP + FN}$ | 0.78 | **0.81** | **0.75** | 0.77 |
| $F1\ score = 2 \cdot \dfrac{Precision \cdot Recall}{Precision + Recall}$ | **0.78** | 0.80 | 0.78 | **0.81** |

Note: All metrics values are between 0 to 1, where 0 indicates the model performance is significantly poor; 1 indicates the model performance is perfect.

| Model | A | B | C | D (Ours) |
|---|---|---|---|---|
| **Input Features** | | | | |
| Site Features | ✓ | ✓ | ✓ | ✓ |
| $CPT_{std}$ and $CPT_{med}$* | | ✓ | | |
| $CPT_{1m\text{-}avg}$** | | | ✓ | |
| $CPT_{Latent}$ | | | | ✓ |
| **Metric** | | | | |
| Accuracy | **0.79** | 0.82 | 0.80 | **0.83** |
| Precision | **0.77** | 0.79 | 0.81 | **0.85** |
| Recall | 0.78 | **0.81** | **0.75** | 0.77 |
| F1 Score | **0.78** | 0.80 | 0.78 | **0.81** |

\* Standard deviation and median values computed over a 4-meter profile below the groundwater table (GWT)
\*\*One-meter interval averages of the CPT profile

To comprehensively evaluate model performance, we use six metrics (see Table 2): accuracy, balanced accuracy, precision, recall, and F1-score, highlighting the highest scores in green and the lowest in red. Model A, trained only on site parameters, performs the worst across all metrics, with an accuracy of 0.79, balanced accuracy of 0.79, and precision of 0.77. This is expected since Model A lacks soil information, which is crucial for assessing liquefaction and lateral spreading.

Model B shows slight improvements, achieving an accuracy of 0.82, balanced accuracy of 0.82, precision of 0.79, recall of 0.81, and an F1-score of 0.80. Model C performs similarly to Model A in most metrics but significantly worse in recall. Although both Model B and C include CPT features as input, the performance of Models B and C indicates that the CPT features based on liquefaction knowledge provide better predictive information than those extracted from averages.

Notably, the proposed Model D provides the best results, with an accuracy of 0.84, precision of 0.85, and the highest F1-score of 0.81. This suggests that even though the autoencoder lacks domain knowledge, the extracted CPT features are useful for predicting lateral spreading. Given that Model D's improvement is due to the extracted CPT features, we will further investigate the impact and representation of these latent features. The following section will employ explainable AI techniques to explore these aspects.

*Explanation of Model D*

In this section, we use SHAP (SHapley Additive exPlanations) to analyze the impact of each feature in Model D. Figure 7 presents the global explanation for Model D. Each dot on the plot represents a data point for a feature, with the x-axis showing the SHAP values and the color indicating the magnitude of the feature values. The features are ranked by importance, and only the top 15 are displayed. The significance of the remaining features is aggregated and presented in the last row. In order to express whether latent features are related to $I_c$ or $q_{c1Ncs}$, we label the latent features as IcX or qcX, where X is for feature numbering; it does not have any physical meaning. For instance, Ic0 means the $I_c$ latent feature indexed with 0.

We observe that four site parameters, elevation, PGA, GWD, and L, are the most important. This indicates that these features strongly predict lateral spreading occurrence, explaining why Model A can achieve comparable performance even without soil information. Beyond site parameters, the most critical latent features are those related to Ic, with Ic3 being the most significant.

Figure 8 illustrates the SHAP values of Ic3 versus its feature values. The x-axis represents the magnitude of Ic3, the y-axis represents the magnitude of SHAP value. In addition, red-ish colors represent data points where GWD is deep, and blue-ish colors represent data points where GWD is shallow. Notably, the high values of Ic3 correspond to high SHAP values, suggesting that high Ic3 values positively impact the risk of lateral spreading, whereas low Ic3 values negatively impact it. Additionally, it is observed that the deep GWD (red-ish color) tends to have a small magnitude of Ic3 SHAP values, implying that deep GWD makes Ic3 have little impact on the prediction.

Although Ic3 is a latent feature without a direct physical interpretation, we can investigate its representation by examining the regions where Ic3 provides positive or negative SHAP values. We select a range of low Ic3 values between -2.3 and -2.5, as most values in this region provide negative SHAP values. Conversely, we select high Ic3 values between +2.4 and +2.5, as most values in this range provide positive SHAP values. We collect data from both regions and reconstruct the $I_c$ profiles. In Figures 9a and 9b, each line represents a reconstructed profile. Figure 9a shows that most reconstructed $I_c$ profiles have low $I_c$ values at depths of 1 to 3 meters, while Figure 9b shows that profiles with Ic3 values between 2.4 and 2.5 have high $I_c$ values at the same depths. These observations align with our understanding of liquefaction: low $I_c$ values indicate sandy soil, which has a higher risk of liquefaction, while high $I_c$ values indicate clay soil.

Figures 9a and 9b suggest that Ic3 is the latent feature influencing the Ic profile pattern at depths of 1 to 3 meters. To further investigate Ic3's impact, we computed the average change in the reconstructed Ic profile compared to a reference profile. We used Ic3 = 0 as the reference point and synthesized 100 data points by sampling the other nine Ic-related latent features from their distributions in the dataset. The reference profile is the average of these 100 reconstructed Ic profiles. We then repeated the process, increasing Ic3 by 0.1, to obtain an average profile for Ic3 = +0.1. The difference between these profiles (ΔIc) represents the influence of an increment in Ic3 (ΔIc3 = +0.1).

In this study, we investigated the impact of ΔIc3 from -4 to +4, with Ic3 = 0 as the reference. The results show that ΔIc3 significantly affects the reconstructed profile at depths of 1-3 meters (see Figure 9c). High ΔIc3 values decrease the Ic value at these depths, while low ΔIc3 values increase the Ic value. This observation aligns with our earlier findings from Figures 9a and 9b: Ic3 control Ic profile pattern at depths of 1 to 3 m. Additionally, the results imply that soil behavior (sand or clay) at 1 - 3 m depths is critical to liquefaction in the Christchurch dataset, as Ic3 is the most important feature among all soil-related features. Moreover, It explains why Ic3 has less impact on the prediction when GWD is deep: Since Ic3 represents the pattern in the shallow layer of Ic profile, it would not contribute to the risk of liquefaction if GWD is deeper.

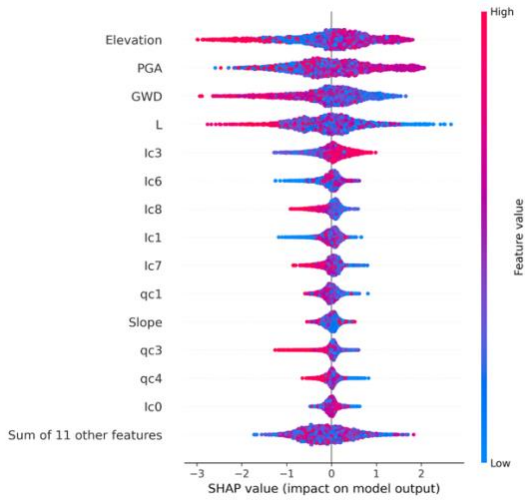 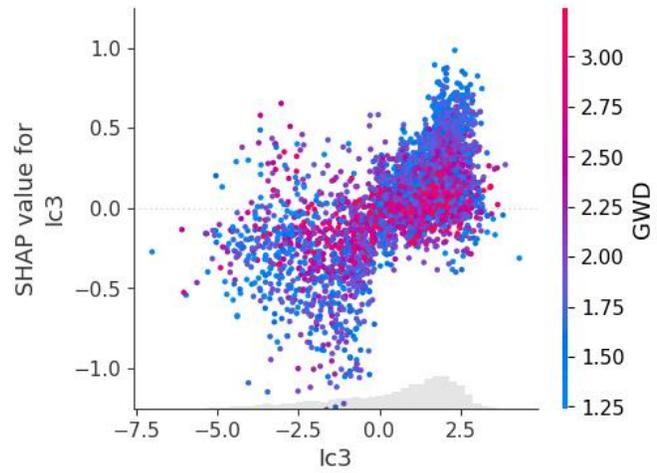

Figure 7. Global Explanation of Model D      Figure 8. Dependency plot of Ic3 versus GWD

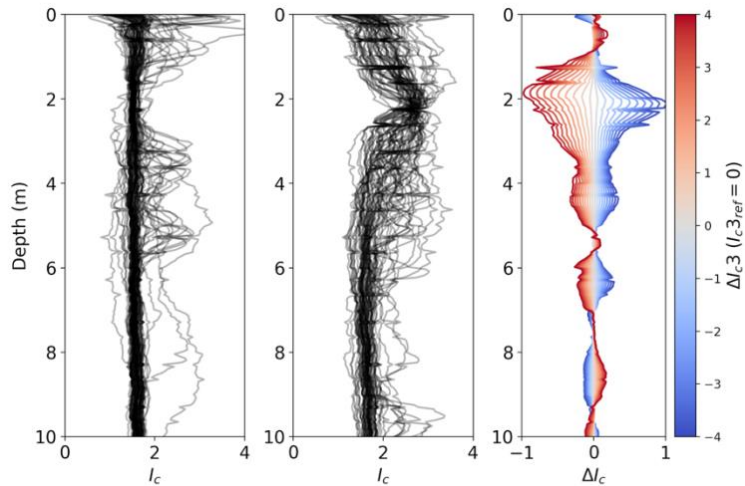

Figure 9. Reconstructed $I_c$ profiles and impact of Ic3 (a) for 2.4<Ic3<2.5 (b) for -2.5<Ic3<-2.3 (c) influence of ΔIc3 on reconstructed $I_c$ profile

## CONCLUSION

In this study, we developed autoencoders to extract latent features from $I_c$ and $q_{c1Ncs}$ profiles, reducing their dimensionality from 200 to 10. The reconstructed results demonstrate that these ten latent features effectively capture the overall trend of the profiles, though some finer details are lost due to the compression. For instance, the autoencoder tends to smooth out abrupt changes in the profile. While increasing the number of extracted features could enhance the detail in the reconstruction, it would contradict the goal of dimensionality reduction.

We assessed the impact of these latent features on predicting lateral spreading. Our results indicate that models using these features achieve the best performance compared to other models, though the improvement is modest.

Using SHAP analysis, we found that latent features related to $I_c$ are generally more important than those related to $q_{c1Ncs}$. Notably, the most critical latent feature, Ic3, influences the pattern of the reconstructed profile at depths of 1 to 3 meters. This finding suggests that in addition to site parameters, soil behavior (whether clay or sand) at a depth of 1 to 3 meters is crucial for predicting liquefaction and lateral spreading.